\pdfoutput=1
\documentclass[11pt]{article}
\usepackage[final]{acl}
\usepackage{stfloats}
\usepackage{times}
\usepackage{latexsym}

\usepackage[T1]{fontenc}
\usepackage[utf8]{inputenc}

\usepackage{microtype}

\usepackage{inconsolata}

\usepackage{graphicx}
\usepackage{amsmath,amsfonts,bm}

\def\eqref#1{equation~\ref{#1}}
\def\1{\bm{1}}

\DeclareMathAlphabet{\mathsfit}{\encodingdefault}{\sfdefault}{m}{sl}
\SetMathAlphabet{\mathsfit}{bold}{\encodingdefault}{\sfdefault}{bx}{n}

\usepackage{listings}
\usepackage{hyperref}
\usepackage{url}
\usepackage{graphicx}
\usepackage{hyperref}
\usepackage{url}
\usepackage{verbatim}
\usepackage{amsthm}
\usepackage{wrapfig}
\usepackage{multirow}
\usepackage{booktabs}
\usepackage{array}%需要该宏包
\usepackage{wrapfig}
\usepackage{bbm}
\usepackage{amssymb}
\usepackage{wrapfig}
\newtheorem{theorem}{Theorem}[section]

\newtheorem{definition}[theorem]{Definition}
\newtheorem{assumption}[theorem]{Assumption}

\def\1{\mathbbm{1}}

\title{Do not Abstain! Identify and Solve the Uncertainty}

\author{%
\textbf{Jingyu Liu}\textsuperscript{1{$\star$}},
\textbf{Jingquan Peng}\textsuperscript{2{$\star$}}, 
\textbf{Xiaopeng Wu}\textsuperscript{2}, 
\textbf{Xubing Li}\textsuperscript{2}, 
\\ % Line break
\textbf{Tiezheng Ge}\textsuperscript{2}, 
\textbf{Bo Zheng}\textsuperscript{2}$^{\dag}$, 
\textbf{Yong Liu}\textsuperscript{1,3,4}$^{\dag}$\\
$^1$ Gaoling School of Artificial Intelligence Renmin University of China Beijing, China \\
$^2$ Taobao \& Tmall Group of Alibaba~ \\
$^3$ Beijing Key Laboratory of Research on Large Models and Intelligent Governance \\
$^4$ Engineering Research Center of Next-Generation Intelligent Search and Recommendation, MOE~ \\
\tt\footnotesize\{liujy1016, liuyonggsai\}@ruc.edu.cn\\
}

\begin{document}
\maketitle
\let\thefootnote\relax\footnotetext{$^\star$ Equal contribution\hspace{3pt} \hspace{5pt}$^{\dag}$ Corresponding author\hspace{5pt}}
\begin{abstract}

Despite the widespread application of Large Language Models (LLMs) across various domains, they frequently exhibit overconfidence when encountering uncertain scenarios, yet existing solutions primarily rely on evasive responses (e.g., "I don't know") overlooks the opportunity of identifying and addressing the uncertainty to generate more satisfactory responses.
To systematically investigate and improve LLMs' ability of recognizing and addressing the source of uncertainty, we introduce \textbf{ConfuseBench}, a benchmark mainly focus on three types of uncertainty: document scarcity, limited capability, and query ambiguity.
Experiments with ConfuseBench reveal that current LLMs struggle to accurately identify the root cause of uncertainty and solve it. They prefer to attribute uncertainty to query ambiguity while overlooking capability limitations, especially for those weaker models. 
To tackle this challenge, we first generate context-aware inquiries that highlight the confusing aspect of the original query. Then we judge the source of uncertainty based on the uniqueness of the inquiry's answer. Further we use an on-policy training method, InteractDPO to generate better inquiries. Experimental results demonstrate the efficacy of our approach. \footnote{code and data in \url{https://github.com/somebodyhh1/ConfuseBench}}

\end{abstract}

\section{Introduction}

Large Language Models (LLMs) \citep{GPT,CAMEL,autogen} have demonstrated remarkable capabilities in a variety of tasks, including text generation, question answering \citep{LLM_powerful,CoT}, code generation \citep{llm_code}, information retrieval \citep{llm_IR} and tool use \citep{toolbench}. However, LLMs tend to exhibit a significant degree of overconfidence \citep{llm_overconfidence,think_twice_overconficence} when faced with question they are not aware of. 

\begin{table}[htbp]
  \centering
  
  \resizebox{7.7cm}{!}{
    \begin{tabular}{cccc}
    \toprule
    Query & Uncertainty & LLM Response & Expectation \\
    \midrule
    unanswerable & low   & I do not know & I do not know \\
    unanswerable & high  & hallucinate & I do not know \\
    answerable  & low   & answer & answer \\
    answerable  & high  & hallucinate & solve it \\
    \bottomrule
    \end{tabular}%
    }
    \caption{Different behavior of LLM when faced with different query. "unanswerable" mean query can not be answered like "weather of 2050."}
  \label{tab:know_unknow}%
\end{table}%

To mitigate this issue, existing researches primarily adopt conservative strategies: response with "I don't know" when identifying potential uncertainties \citep{knowledge_of_knowledge,no_I_dont_know,know_unknow,llm_know_when_not_answer}. However, this strategy exhibits significant limitations. As shown in Table \ref{tab:know_unknow}, for inherently unknowable questions (e.g., "weather of 2050."), models should consistently response with "I don't know". However, for those answerable queries, simply response with "I do not know" overlooks the opportunity of addressing the uncertainty, failing to generate more satisfactory responses. Specifically, when confidence levels are low (e.g., "quantum computing's impact on climate modeling"), the system should proactively identify uncertainty sources (insufficient document/reasoning capability gap/query ambiguity), then employ dynamic strategies such as retrieval \citep{RAG,RAG_RAFT,lin_RAG}, CoT \citep{CoT,COT_serial}, or clarification \citep{tell_me_more,MAQA} to improve the response quality.

To investigate and improve LLMs' performance on identifying and solving the uncertainty, we introduce \textbf{ConfuseBench}, a benchmark that encompasses three distinct types of uncertainty: document scarcity, limited capacity, and query ambiguity. Document scarcity occurs when models lack essential factual information to answer a question, and additional documents could provide assistance \citep{RAG,RAG_RAFT,RAG_reasoning}; Limited capacity indicates that the query is too complex for the model to resolve effectively, in such cases, a larger model or extended reasoning steps might be beneficial \citep{CoT,ToT}. Query ambiguity occur when the query itself is unclear, where multiple answers may suffice or the query may not be answerable at all, necessitating further clarification \citep{ambigqa,tell_me_more,clamber}. Through ConfuseBench, we surpass conventional evaluation paradigms that focus solely on answer accuracy or basic uncertainty detection. Instead, ConfuseBench rigorously assesses models' capacity to (1) diagnose the root causes of uncertainty and (2) actively mitigate such uncertainty to generate substantively improved responses.

Our paper mainly consider the situation where the model can not directly tell the answer of the question due to uncertainty and there is some method to help solving the uncertainty and generate a better response. We consider three main methods to improve the quality of generation: retrieval, CoT and clarification, which corresponds to three different source of uncertainty: document scarcity, limited capacity and query ambiguity. Those three methods to improve the quality should be most common ones so the three source of uncertainty also incorporate most situations. We show some examples in Appendix \ref{app:fewshotexample}

Our experiments with ConfuseBench have revealed that current models including GPT-4o struggle to identify the sources of uncertainty, which leads to unsatisfying performance on this benchmark. Those models prefers to categorize questions as ambiguous and request the user for clarification. For example, when we provide the model a clear query "locate the best yoga class in New York" and a noise document about yoga classes in London, the model might regard the query as ambiguous and response with "Are you referring to London?". Furthermore, the models seldomly acknowledge failures caused by their own capability limitations, when confronted with uncertainty, models often attribute the issue to external factors rather than recognizing their own limitations.

To address this issue, we propose a two-step approach. Instead of directly identifying the source of uncertainty, we first focus on accurately locating the confusing parts of the problem and generating a follow-up inquiry. If the answer to inquiry is an objective fact, the retrieval system can effectively provide the required information. If multiple answers fit the inquiry appropriately, further clarification is necessary. Conversely, if the follow-up inquiry is logically incoherent or merely paraphrased repetitions of the original question, it means the model fails to effectively understand the query and CoT could be beneficial. Furthermore, to enhance the capability of generating effective follow-up inquiry, we propose the InteractDPO, a training paradigm that dynamically generates "chosen-rejected" sample pairs through real-time interaction with retrieval systems or users during training, thereby achieving on-policy optimization.
    
\begin{figure}
    \centering
    \includegraphics[width=1\linewidth]{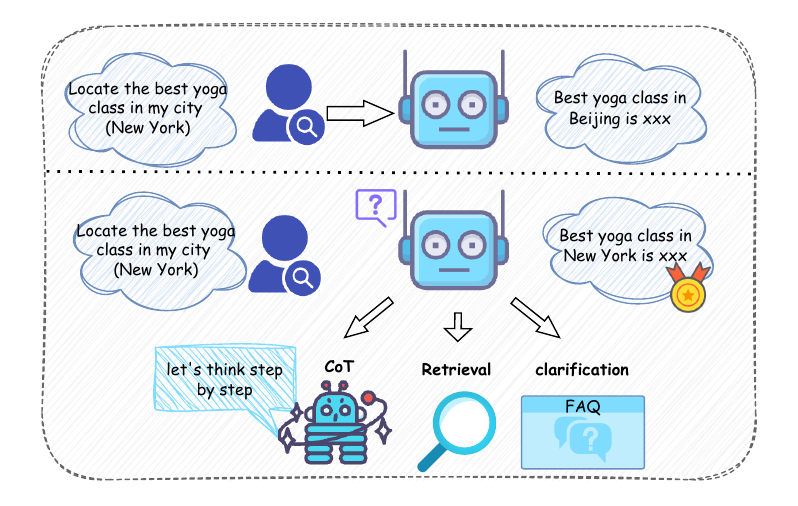}
    \caption{LLMs recognize different source of uncertainty and try to solve the uncertainty.}
    \label{fig:illustration}
\end{figure}

Overall, our key contributions include:
1) This paper introduces a new benchmark designed to measure LLMs' ability to identify different types of uncertainty arising from various sources, including document scarcity, limited capability, and query ambiguity. 2) We demonstrate through experiments that current LLMs exhibit significant challenges in reliably differentiating between these sources of uncertainty. 3) We propose a novel method for identifying the source of uncertainty based on the uniqueness of the inquiry's answer, and we further enhance the inquiry generation process through InteractDPO.

\section{Related Work}

\textbf{Recognizing the Uncertainty.} \citet{knowledge_of_knowledge,do_llm_know_they_dont} propose that models should learn to understand what they do not know instead of giving a hallucinated answer. \citet{llm_unknow_prompt,knowledge_of_knowledge,llm_know_when_not_answer} further design various of prompts to instruct LLM express "I do not know" when encountering uncertainty. \citet{alignment_honesty,llm_express_uncertainty} try to finetune the model to express how uncertain it is in verbal language and \citet{no_I_dont_know} propose to train the model to give some explanation to the unanswerability. \citet{no_I_dont_know,clamber} also categorize why a question is unknown, but they mainly focus on ill-defined input, ignoring lack of capacity and documents, they still fail to recognize and solve the source of uncertainty.

\noindent \textbf{Solving the Uncertainty.} For knowledge based uncertainty, \citet{iterative_retrieval,llms_know_what_need_RAG} try to solve multi-hop queries by iterative reasoning and retrieving until the model feels confident enough to provide an answer, \citet{Adaptive_RAG,adaptive_note,efficientRAG} also iteratively call the model to generate retrieval query to solve the uncertainty. For ambiguity based uncertainty, \citet{tell_me_more} construct Intention-in-Interaction (IN3) to evaluate the ability of asking clarification question, \citet{learning_to_ask} prompts the LLM to adaptively ask clarification questions and \citet{MAQA} further propose to use prompt, entropy and logits to measure is clarification needed. However, these works are constrained to only one source of uncertainty, fail to consider the situation that the uncertainty may rise from other sources.

\noindent \textbf{Uncertainty Decomposition.}  As uncertainty could be raised from different sources, recognizing the source is an important topic \citep{uncertainty_3,uncertainty_survey,uncertainty_survey_2}, previous works typical classify uncertainty into data and model uncertainty. \citet{decompose_clarification} try to judge does the query needs to be clarified by observing how will the model perform when faced with different clarifications. \citet{decopose_context} use ensemble methods and use different in context example to simulate models to decompose the uncertainty. \citet{To_believe} propose a new definition of model uncertainty and decompose uncertainty by distribution shift when some answers are provided to the LLM. But those methods simply classifies the uncertainty as data uncertainty and model uncertainty, fails to consider the real uncertainty types the LLM would met in application.

\section{Benchmark Construction}

Previous benchmarks have primarily focused on refusing to answer unknown queries \citep{knowledge_of_knowledge,no_I_dont_know}, or have merely considered iterative retrieval and clarification techniques \citep{adaptive_note,efficientRAG,tell_me_more}. This approach fails to recognize that models need to identify the source of uncertainty and implement corresponding measures to address it.
To comprehensively enhance and quantitatively evaluate these capabilities in model designs, we introduce \textbf{ConfuseBench}, a benchmark that encompasses various sources of uncertainty. This benchmark aims to assess and inspire LLMs' abilities to recognize and resolve uncertainties effectively.

We evaluate three main scenarios in which LLMs are commonly employed: basic question answering, assistant interactions, and tool utilization. To assess the ability of recognizing and resolving uncertainty, we have collected various datasets and rewritten queries and associated documents to create the case holding certain uncertainty. For basic question answering, we incorporate HotpotQA\citep{hotpotqa} and AmbigQA \citep{ambigqa}. In the assistant scenario, we consider ExpertQA \citep{expertQA} and TechQA \citep{techQA}. We utilize ToolBench \citep{toolbench} for tool usage. 
It is worth noting that, to facilitate this evaluation, we employ GPT-4o to generate the tool calling chain for ToolBench, using the calling chain as the answer rather than the actual calling result.

To construct data cases where uncertainty arises from insufficient capability, we instruct the Large Language Model (LLM) to generate answers based on query and gold documents. If the model fails to produce the correct answer (which is already indicated in the documents), we attribute the uncertainty to its insufficient capability. Conversely, if the model successfully generates the correct answer, we construct a new document set by randomly discarding portions of the gold documents and retrieve some new ones. If the model cannot produce the correct answer under the new document set, we classify the uncertainty as stemming from missing documents.
It is important to note that different large language models possess varying knowledge and capability boundaries. During evaluation, if a model can generate correct answers based on the original query and provided documents, it is deemed free of uncertainty, and such cases will be excluded from the evaluation.

    \begin{figure}
        \centering
        \includegraphics[width=1\linewidth]{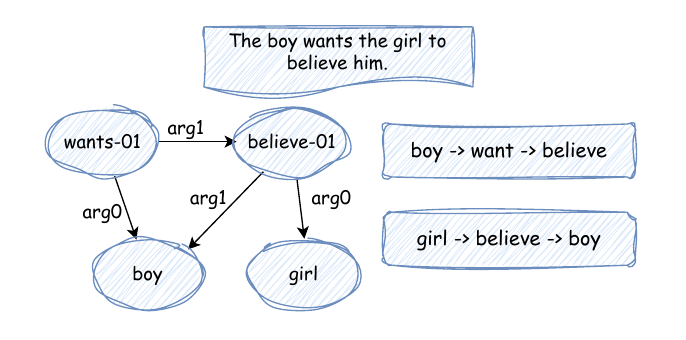}
        \caption{Abstract Meaning Representation for "The boy wants the girl to believe him."}
        \label{fig:AMR}
    \end{figure}

    For uncertainty arising from ambiguity, we directly utilize the ambiguous queries provided in AmbigQA \citep{ambigqa}. For the other four datasets, we first transform the queries into Abstract Meaning Representation (AMR) \citep{AMR}, where each query is represented by entity nodes and the corresponding relationships between those entities, forming a graph-based structure as shown in Figure \ref{fig:AMR}. 
    Subsequently, we prompt GPT-4o to introduce ambiguity into the AMR graph by removing modifiers and descriptive words, omitting key information, altering the relationships between nodes, and reorganizing the AMR structure. This method enables the model to better understand the semantic structure of the query, allowing us to provide clearer and more direct instructions for transforming the AMR into an ambiguous query.
    Then, we convert the AMR into an ambiguous query and generate the corresponding clarifications. If the model fails to answer the ambiguous query but successfully responds to it when provided with the clarification, we categorize the query as ambiguous.

\begin{table}[htbp]
  \centering
  
  \resizebox{\linewidth}{!}{
    \begin{tabular}{c|cccc}
    \toprule
          &       & document & ambiguity & ability \\
    \midrule
    \multirow{2}[2]{*}{QA} & HotpotQA & 859   & 702   & 141 \\
          & AmbigQA & 543   & 537   & 167 \\
    \midrule
    \multirow{2}[2]{*}{Assistant} & ExpertQA & 442   & 397   & 141 \\
          & TechQA & 470   & 683   & 140 \\
    \midrule
    Tool Usage & ToolBench & 479   & 590   & 144 \\
    \bottomrule
    \end{tabular}
    }%
    \caption{statistics of the benchmark}
    \label{tab:benchmark_statistic}%
\end{table}%

The statistics of the dataset is shown in Table \ref{tab:benchmark_statistic}. Additionally, we manually select 50 cases each for queries lacking documentation and those that are ambiguous, as well as 30 cases for instances categorized as lacking ability, from each dataset to construct the benchmark. The remaining cases are used as training data. Consequently, the benchmark comprises a total of \(5 \times (50 + 50 + 30) = 650\) cases.

\section{Preliminary Test}
\begin{table*}[htbp]
  \centering

    \begin{tabular}{cccccccc}
    \toprule
          &       & HotpotQA & AmbigQA & TechQA & ExpertQA & ToolBench & avg \\
    \midrule
    \multirow{6}[2]{*}{AQ} & GPT-4o & 0.446  & 0.600  & 0.760  & 0.729  & 0.846  & 0.6762  \\
          & DeepSeek-V3 & 0.515  & 0.600  & 0.783  & 0.779  & 0.794  & 0.6942  \\
          & Qwen2.5-72B & 0.585  & 0.631  & 0.804  & 0.763  & 0.812  & 0.7188  \\
          & Llama-3-70B & 0.438  & 0.538  & 0.758  & 0.721  & 0.725  & 0.6362  \\
          & Qwen2.5-7B & 0.369  & 0.500  & 0.742  & 0.740  & 0.744  & 0.6181  \\
          & Mistral-7B & 0.400  & 0.492  & 0.733  & 0.735  & 0.794  & 0.6308  \\
    \midrule
    \multirow{6}[2]{*}{UCA} & GPT-4o & 0.531  & 0.377  & 0.477  & 0.400  & 0.685  & 0.4938  \\
          & DeepSeek-V3 & 0.462  & 0.431  & 0.400  & 0.438  & 0.562  & 0.4585  \\
          & Qwen2.5-72B & 0.631  & 0.592  & 0.431  & 0.408  & 0.700  & 0.5523  \\
          & Llama-3-70B & 0.438  & 0.408  & 0.408  & 0.385  & 0.400  & 0.4077  \\
          & Qwen2.5-7B & 0.431  & 0.454  & 0.415  & 0.408  & 0.415  & 0.4246  \\
          & Mistral-7B & 0.438  & 0.415  & 0.438  & 0.446  & 0.531  & 0.4538  \\
    \bottomrule
    \end{tabular}%
  \caption{Performance of locating and solving the uncertainty. AQ represents the quality of answer after interaction; UCA is the uncertainty classification accuracy.}
  \label{tab:prompt_answer_performance}%
\end{table*}%

To evaluate the ability to address uncertainty, we instruct the LLM to determine whether it should interact with the retrieval system, consult a user, or utilize Chain of Thought (CoT) reasoning to resolve the uncertainty. If the model opts for CoT, it will generate an answer through a chain of thought reasoning. Conversely, if it chooses to engage with the retrieval system, it will generate a query to retrieve additional documents and then answer the original question based on the results of the interaction. If the model decides to interact with a user, it will formulate inquiries to ask clarifying questions and provide answers based on the received clarifications. We use GPT-4o to simulate the user and provide clarifications based on the inquiries made by the model. And we use majority voting when judging the source of uncertainty, specifically, we generate for 3 times, if the first two matches, then it is considered as the answer, otherwise we use the third generation.

We primarily evaluate the following metrics:
\begin{itemize}
    \item \textbf{Answer Quality(AQ)}: This metric assesses the quality of the answer provided after interaction or using Chain of Thought (CoT). For the HotpotQA and AmbigQA datasets, we employ Qwen2.5-72b as a judge to evaluate correctness. For the other datasets, we score answers based on their usefulness on a scale from 1 to 4; the results shown below are normalized to a range of 0-1.
    \item \textbf{Uncertainty Classification Accuracy(UCA)}: This measures the LLM's capacity to recognize the source of uncertainty, knowing that it should interact with the retrieval system, the user or solve it by CoT.
    \item \textbf{Inquiry Quality(IQ)}: This metric evaluates the quality of the inquiries generated. We compare the query before ambiguity and the gold documents with the actual query and documents provided to the model to derive a gold standard inquiry. We then assess how closely the actual inquiry aligns with the gold inquiry and score it on a scale from 1 to 4 and we normalize it to 0-1 in the paper.
\end{itemize}

We mainly assess the following models: GPT-4o-0513, DeepSeek-V3, Qwen2.5-72b-Instruct, Meta-Llama-3-70B-Instruct, Qwen2.5-7b-Instruct, and Mistral-7B-Instruct-v0.2. We use these models to judge the source of uncertainty and generate corresponding inquiries. We aim to evaluate the ability of locating and solving the uncertainty rather than the ability of solving the problem, therefore, to avoid the impact of different perceptions of the question by the models themselves, we use both the evaluated model and GPT-4o to generate answers based on the interaction results, the highest score is considered. To measure the reliability LLM evaluating the performance, we show the omega reliability in Appendix \ref{app:reliability}

From Table \ref{tab:prompt_answer_performance}, we can observe that the LLM fails to effectively recognize the source of uncertainty and generate corresponding inquiry to solve the uncertainty. The best performing model only successfully classify about 50\% of cases. Those weaker models like Mistral-7b and Qwen2.5-7b fails to effectively recognize the source of uncertainty, even Llama-3-70B shows unsatisfying performance.

\begin{table}[htbp]
  \centering
  
  \resizebox{7.7cm}{!}{
    \begin{tabular}{ccccc}
    \toprule
          & metric & doc   & ambig & ability \\
    \midrule
    \multirow{2}[2]{*}{GPT-4o} & precision & 0.53  & 0.41  & 0.28 \\
          & recall & 0.19  & 0.76  & 0.2 \\
    \midrule
    \multirow{2}[2]{*}{Llama-3-70B} & precision & 0.3   & 0.39  & 0.22 \\
          & recall & 0.12  & 0.97  & 0.02 \\
    \midrule
    \multirow{2}[2]{*}{Qwen2.5-7B} & precision & 0.43  & 0.39  & 0.31 \\
          & recall & 0.1   & 0.94  & 0.1 \\
    \midrule
    \multirow{2}[2]{*}{Mistral-7B} & precision & 0.44  & 0.41  & 0.33 \\
          & recall & 0.12  & 0.85  & 0.14 \\
    \bottomrule
    \end{tabular}%
    }
    \caption{Precision and recall of different uncertainty}
  \label{tab:precision_recall}%
  
\end{table}%

\begin{figure}[h]
    \centering    
    \includegraphics[width=\linewidth]{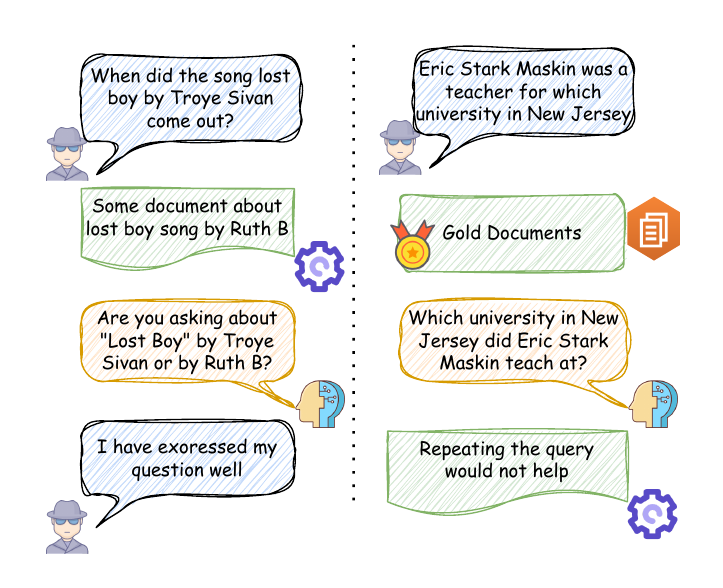}
    \caption{In the left case, the model retrieved some documents about another singer and asks the user to change the query. In the right case, the model simply rephrase the query and wants to retrieve more information.}
    \label{fig:case_study}
\end{figure}

From Table \ref{tab:precision_recall}, we can observe that when faced with uncertainty, LLMs tend to attribute uncertainty to query ambiguity ("ambig") rather than insufficient document support ("doc"), particularly in the less powerful models as indicated by the high recall of "ambig" and low recall of "doc". 

We can observe that the classification is unbalanced; they recognizes most of the queries as ambiguous and proceeds to interact with the user instead of more retrieval and lack of capacity.

As illustrated in Figure \ref{fig:case_study}, when presented with a clear query accompanied by noisy documents, the model can become distracted by the noise. It may ask the user to clarify the query, hoping to change the intention of the user so that it can leverage information from the noisy documents to generate an answer. 
In this context, the model understands why it cannot answer the query, but it places greater trust in the relevance of the documents than in the clarity of the query. Consequently, instead of seeking supplemental documents, the model attempts to align user intention with available information in the noisy documents, ultimately resulting in the unexpected behavior of requiring query rephrasing rather than acquiring more relevant documents.

Moreover, large language models (LLMs) seldom acknowledge that they cannot answer a question due to a lack of capability, as shown in Table \ref{tab:precision_recall}. In our view, this phenomenon is similar to overconfidence \citep{llm_overconfidence,think_twice_overconficence,llm_express_uncertainty}; when faced with uncertainty, these models often provide incorrect answers instead of recognizing their limitations with a response such as, "I don't know." And when instructed to choose the reason for their uncertainty, they also tend to refuse to acknowledge that it is due to their limited reasoning capacity and blame it to insufficient documents or ambiguity.

As we show in Figure \ref{fig:case_study}, when faced with uncertainty brought by capacity, the model might direct rephrase the query, this can be explained in two ways: 1) the model fails to recognize any lack of documents or ambiguity, so it can only repeat the query again; 2) what confuses the model is the query itself, it fails to effectively understand the query and the given documents, so the query itself is the confusing part.

\begin{figure}
    \centering
    \includegraphics[width=1\linewidth]{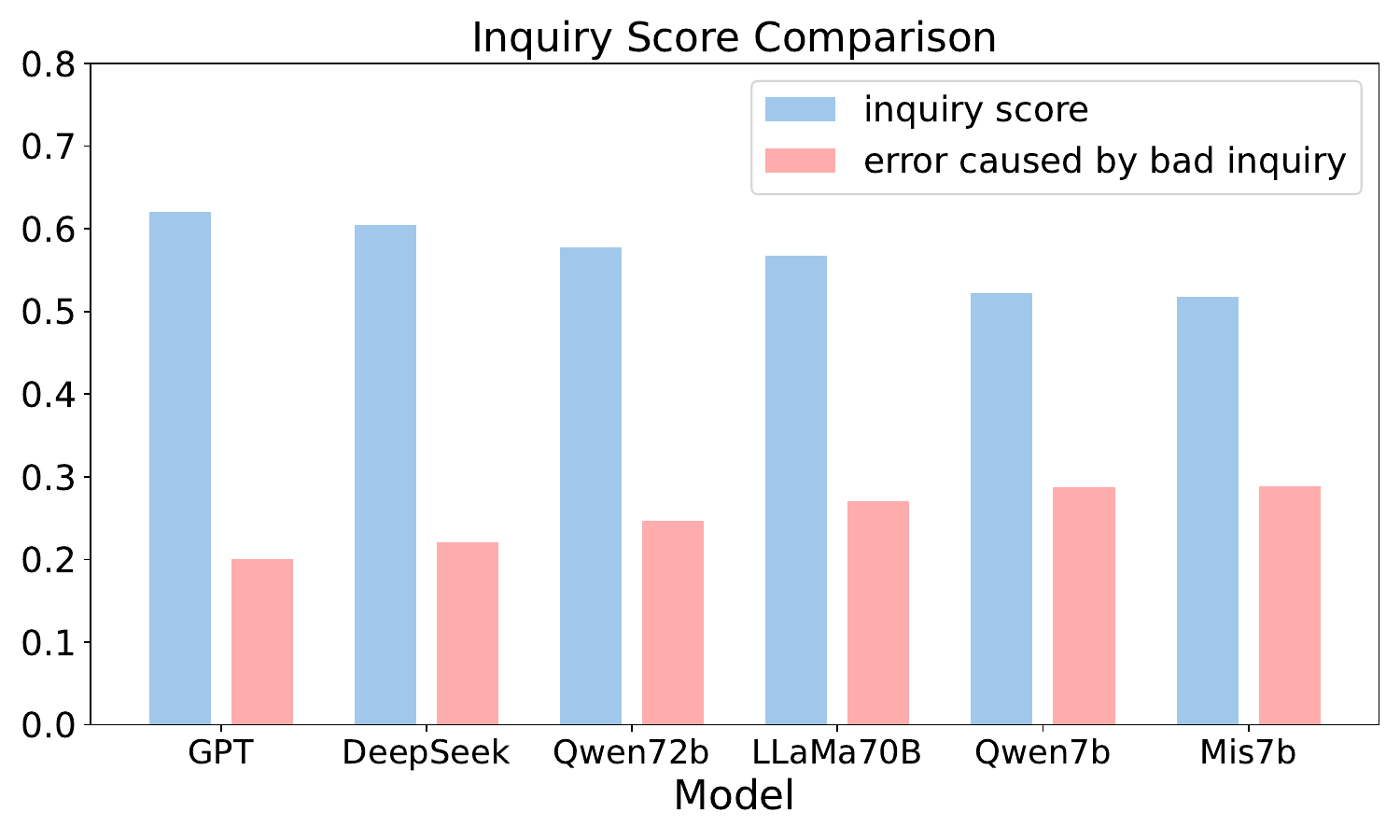}
    \caption{The inquiry score and the percentage of errors caused by bad inquiry (correctly classified but answer incorrectly with low inquiry score)}
    \label{fig:inquiry_evaluation}
\end{figure}

For the quality of inquiry, we can observe from Figure \ref{fig:inquiry_evaluation} that powerful models like GPT-4o are capable of generating meaningful inquiries, which could help to gather some useful information, but the quality of inquiry could still be improved and smaller models such as Qwen2.5-7b, performs worse.

\section{Method}

In this section, we begin by discussing the use of CoT to identify factors that may be confusing the model and to assess the sources of uncertainty. We then propose that the uncertainty associated with the inquiry is equivalent to that of the query itself. This allows us to directly evaluate the uncertainty by examining the inquiry. And we propose to utilize the uniqueness of the inquiry answer to recognize the source of uncertainty.

\subsection{Judge Based on Inquiry Answer}\label{sec:inquiry_class}

Apparently, judging the source of uncertainty is a difficult task, and less powerful models fail to complete this task; they prefer to regard the query as ambiguous and interact with the user. However, we can also observe from Figure \ref{fig:inquiry_evaluation} that these models can effectively generate inquiries to interact with their environment, with inquiry scores not significantly lower than GPT-4o and DeepSeek-V3 \citep{ask_when_a2b2,tell_me_more}. A natural approach is to leverage Chain of Thought to identify what is confusing the model, and judge the source of uncertainty afterwards.

Also, we can show that the uncertainty held by the inquiry is actually the same with the query. And the inquiry typically only involve some sub-aspects of the query, it should be easier to identify the source of uncertainty based on the inquiry.

\begin{definition}
    Consider a query $x$, the corresponding answer $y$, document $d$ and the clarification to the query $c$. Let $\theta$ be the model and $\theta^*$ be the optimal model which can perfectly solve the query $x$. Then, the uncertainty raised by capacity $U_c$, by knowledge $U_k$ and by ambiguity $U_a$
    \begin{equation}
        \begin{split}
            U_c&=H(y|x,d,c,\theta),\\
            U_k&=H(y|x,c,\theta^*),\\
            U_a&=H(y|x,d,\theta^*),
        \end{split}
    \end{equation}
    where $H(\cdot)$ stands for entropy
\end{definition}
Therefore, $U_c$ is the uncertainty of the model when all information is given, so it is raised by lack of capacity. $U_{k}$ and $U_{a}$ is the uncertainty for the optimal model when documents and clarification are missing, they correspond to uncertainty raised by lack of documents and ambiguity. Then, we can show that the inquiry holds similar uncertainty with the query.

\begin{theorem}
    Given a query $x$ and the generated inquiry $q$, then the uncertainty of $q$ is positively related to the uncertainty of $x$, then,
    \begin{equation}
    \begin{split}
        |U_{k}(q)-U_{k}(x)|  &\leq -\log p(q^*|x,c,\theta),\\
        |U_{a}(q)-U_{a}(x)|  &\leq -\log p(q^*|x,d,\theta),\\
        |U_{c}(q)-U_{c}(x)| & \leq  -\log p(q^*|x,d,c,\theta),\\
    \end{split}
        \nonumber
    \end{equation}
    where $q^*$ is the optimal inquiry generated by $\theta ^*$. For lack of ability, the optimal inquiry is the original query.
\end{theorem}

The theorem posits that if the model generates a meaningful inquiry, the uncertainty held by the inquiry is similar to the original query. Otherwise if the inquiry is meaningless, it shows that the model fails to understand the query well, and it also shows lack of capacity.

Therefore, we can enhance the performance based on the generated inquiry. If the generated inquiry fails to recognize the confusing part, then we classify it as a lack of capacity, indicating that Chain of Thought is required. Conversely, if the inquiry requires additional documents to solve, retrieval is required otherwise clarification is needed.

    \begin{figure}
        \centering
        \includegraphics[width=1\linewidth]{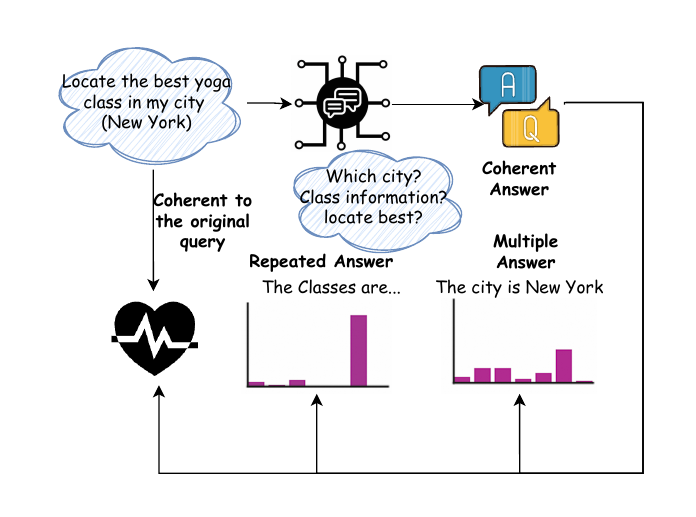}
        \caption{Judge the source of uncertainty based on answer of inquiry}
        \label{fig:judge_by_answer}
    \end{figure}

But when facing some complex query, directly judge the source of uncertainty based on the inquiry is hard.
To measure the uncertainty held by inquiry, we propose utilizing the answer of inquiry to help the judgment when facing complex problems. First, if the uncertainty arises from a lack of capability, the model would merely rephrase the query; thus, the response to the inquiry should appropriately address the original query. Consequently, we can determine whether the uncertainty stems from a lack of capability by evaluating the semantic coherence when the inquiry answer is considered as the answer to the original query. 
In this way, we can not only recognize cases where Chain of Thought (CoT) is needed but also prevent unnecessary retrieval and clarification.

Also, if the inquiry requires extra retrieval, it typically indicates that the answer points to a definitive objective fact. Conversely, if the query needs further clarification, it suggests the question may have multiple valid subjective answers. To distinguish these scenarios, we designed a verification method inspired by \citet{To_believe}: First, we provide the LLM with a logically coherent preset answer to the inquiry. We then instruct the model to generate a new distinct response based on this input. For objective factual questions, the model—lacking prior knowledge—tends to directly repeat the fabricated answer. However, for open-ended subjective questions, the model recognizes the potential for diverse solutions and can still produce novel, reasonable responses even after receiving the preset answer.

For example, consider the query \textit{Locate the best yoga class in my city} and the corresponding inquiry \textit{Which city are you referring to?} If "New York" is provided as a possible answer, the model can easily generate an alternative answer like "London." However, if the inquiry is \textit{Find the yoga classes in New York} and an answer is provided, the model is likely to repeat that answer, indicating it does not know any other answers. Therefore, when it is hard for the model to directly judge the source of uncertainty, judge based on answer could help. So we first judge the source based on inquiry twice, and if the two samples differ, we judge the source of uncertainty by answer.

\subsection{Inquiry Quality Matters}

It is important to note that if the model fails to generate an inquiry of high quality, the advantages of a more concise input may be overshadowed by the drawbacks of a poor inquiry, resulting in suboptimal performance, and a good inquiry can help the model to gather more useful information. Therefore, enhancing the quality of the generated inquiry is essential.

\begin{table*}[h]
  \centering

    \begin{tabular}{cccccccc}
    \toprule
          &       & HotpotQA & AmbigQA & TechQA & ExpertQA & ToolBench & avg \\
    \midrule
    \multirow{3}[2]{*}{GPT-4o} & prompt & 0.531  & 0.377  & 0.477  & 0.400  & 0.685  & 0.4938  \\
          & inquiry & 0.523  & 0.562  & 0.538  & 0.469  & 0.754  & 0.5692  \\
          & answer & \textbf{0.600 } & \textbf{0.585 } & \textbf{0.562 } & \textbf{0.515 } & \textbf{0.769 } & \textbf{0.6062 } \\
    \midrule
    \multirow{3}[2]{*}{DeepSeek-V3} & prompt & 0.462  & 0.431  & 0.400  & 0.438  & 0.562  & 0.4585  \\
          & inquiry & \textbf{0.546 } & 0.538  & 0.446  & 0.492  & 0.662  & 0.5369  \\
          & answer & 0.546  & \textbf{0.562 } & \textbf{0.485 } & \textbf{0.508 } & \textbf{0.669 } & \textbf{0.5538 } \\
    \midrule
    \multirow{3}[2]{*}{Qwen2.5-72B} & prompt & \textbf{0.631 } & 0.592  & 0.431  & 0.408  & 0.700  & 0.5523  \\
          & inquiry & 0.592  & 0.631  & 0.554  & 0.400  & 0.708  & 0.5769  \\
          & answer & 0.608  & \textbf{0.654 } & \textbf{0.569 } & \textbf{0.431 } & \textbf{0.754 } & \textbf{0.6031 } \\
    \midrule
    \multirow{3}[2]{*}{Llama-3-70B} & prompt & 0.438  & 0.408  & 0.408  & 0.385  & 0.400  & 0.4077  \\
          & inquiry & \textbf{0.631 } & 0.469  & \textbf{0.538 } & 0.515  & 0.531  & 0.5369  \\
          & answer & 0.615  & \textbf{0.492 } & 0.515  & \textbf{0.515 } & \textbf{0.600 } & \textbf{0.5477 } \\
    \midrule
    \multirow{3}[2]{*}{Qwen2.5-7B} & prompt & 0.431  & 0.454  & 0.415  & 0.408  & 0.415  & 0.4246  \\
          & inquiry & 0.462  & 0.477  & 0.523  & 0.469  & 0.454  & 0.4769  \\
          & answer & \textbf{0.500 } & \textbf{0.531 } & \textbf{0.523 } & \textbf{0.485 } & \textbf{0.462 } & \textbf{0.5000 } \\
    \midrule
    \multirow{3}[2]{*}{Mistral-7B} & prompt & 0.438  & 0.415  & 0.438  & 0.446  & 0.531  & 0.4538  \\
          & inquiry & 0.508  & 0.562  & 0.546  & 0.515  & 0.515  & 0.5292  \\
          & answer & \textbf{0.515 } & \textbf{0.562 } & \textbf{0.562 } & \textbf{0.546 } & \textbf{0.554 } & \textbf{0.5477 } \\
    \bottomrule
    \end{tabular}%
    \caption{Uncertainty classification performance of using CoT to judge uncertainty after inquiry generation (inquiry) and judge the uncertainty by inquiry answer (answer), prompt means directly judge the uncertainty source.}
  \label{tab:per_by_answer}%
  
\end{table*}%

\begin{table*}[h]
  \centering
 
    \begin{tabular}{ccccccc}
    \toprule
          & HotpotQA & AmbigQA & TechQA & ExpertQA & ToolBench & avg \\
    \midrule
    GPT-4o & 0.600  & 0.585  & \textbf{0.562 } & \textbf{0.515 } & 0.769  & 0.6062  \\
    vanilla & 0.554  & 0.538  & 0.454  & 0.438  & 0.731  & 0.5431  \\
    SFT   & 0.585  & 0.569  & 0.485  & 0.485  & 0.746  & 0.5738  \\
    DPO   & 0.608  & 0.554  & 0.515  & 0.485  & 0.762  & 0.5846  \\
    \midrule
    onlineDPO & 0.608  & 0.569  & 0.531  & 0.492  & 0.762  & 0.5923  \\
    InteractDPO & \textbf{0.615 } & \textbf{0.592 } & 0.546  & 0.508  & \textbf{0.769 } & 0.6062  \\
    \bottomrule
    \end{tabular}%
     \caption{Uncertainty classification performance of InteractDPO}
  \label{tab:per_interact}%
\end{table*}%

Therefore, we propose \textbf{InteractDPO}. Vanilla DPO use preference datasets collected ahead of training, the responses in the dataset are usually generated by different LLMs \citep{DPO,onlineDPO}. Thus, the feedback is usually purely offline. To conduct on-policy training, we first collect some preference datasets, then during training, the trained model generates an inquiry based on the prompt and interact with the retrieval system or the user-GPT to gather more documents or clarification. The model then generates answer based on the interaction. During training, if the trained model successfully generate an inquiry to solve the original query, it will be selected as the chosen inquiry, otherwise the rejected inquiry to conduct on policy DPO training. OnlineDPO \citep{onlineDPO} instead generate two different inquiry and use the LLM to judge which one is better and mark is as chosen, the other one as rejected. But the LLM can not accurately judge which one is better without real supervision signal, therefore, compared to directly use LLM to select the chosen-rejected pair, InteractDPO provides real feedback and shows better performance when classifying the source of uncertainty and gathering information.

\section{Experiments}

To validate the performance of our proposed method, we conduct experiments on the benchmark. When judging the source of uncertainty, we sample three times and use majority voting to judge source of uncertainty, we also use prompt judge is it lack of capacity for all 3 methods due to its high precision. We repeat the experiment for 3 times and use the averaged performance.

As shown in table \ref{tab:per_by_answer}, judging by the inquiry and the answer can help to increase the performance, directly judging based on the inquiry can help the model to better judge the source of uncertainty, results of more models, the answer quality are shown in Appendix \ref{app:further_experiments}. And the results of few shot are shown \ref{app:fewshotexample}.

For InteractDPO, based on Figure \ref{fig:inquiry_evaluation}, Qwen2.5-7B show great performance when generating inquiry, therefore, we choose the model to conduct further finetuning and enhance its ability of generating high quality inquiries. As we mainly want to enhance and evaluation of generating inquiry, we use the finetuned model to generate inquiry, and use GPT-4o to conduct classification and further answering. Also, we compare our method with SFT, DPO and OnlineDPO, as shown in Table \ref{tab:per_interact}, InteractDPO helps the most. The performance of answer quality is shown in Appendix \ref{app:further_experiments}.

\section{Conclusion}
In this paper, we discuss the fact that current LLMs fail to effectively judge the source of uncertainty. 
Models prefers to recognize the query as ambiguous seldomly admit lack of capacity. Then, we propose to judge the source of uncertainty by uniqueness of inquiry answer, to further increase the performance, we propose InteractDPO to help the model generate better inquiry.

%By constructing a benchmark containing three main scenarios of LLM usage, we find that models prefers to recognize the query as ambiguous and ask for some clarification, and the model seldomly admit that it would fail the task simply because lack of capacity. Then to solve this problem, we propose to judge the source of uncertainty by inquiry and the uniqueness answer of inquiry, then to further increase the performance, we propose InteractDPO to help the model generate better inquiry.

\section*{Limitations}
This paper primarily discusses how current large language models (LLMs) fail to recognize the sources of uncertainty. While we focus on three main categories of uncertainty, these can be further specified. For instance, a lack of documents may correspond to deficiencies in factual knowledge or background information, each requiring different databases for retrieval. Regarding lack of ability, while Chain of Thought (CoT) techniques can address some issues, there are also cases that necessitate the use of methods like Tree of Thought or Monte Carlo Tree Search (MCTS). And there is also various of reasons why the query is ill-defined for example, it could be ambiguous or factually incorrectly or asks for a illegal time. Consequently, there are various sources of uncertainty, each linked to its own solution; however, we only examine the three most common ones.

\section*{Acknowledgements}
This research was supported by National Natural Science Foundation of China(No.62476277), National Key Research and Development Program of China(NO. 2024YFE0203200), CCF-ALIMAMA TECH Kangaroo Fund(No.CCF-ALIMAMA OF 2024008), and Huawei-Renmin University joint program on Information Retrieval. We also acknowledge the support provided by the fund for building worldclass universities (disciplines) of Renmin University of China and by the funds from Beijing Key Laboratory of Big Data Management and Analysis Methods, Gaoling School of Artificial Intelligence, Renmin University of China, from Engineering Research Center of Next-Generation Intelligent Search and Recommendation, Ministry of Education, from Intelligent Social Governance Interdisciplinary Platform, Major Innovation \& Planning Interdisciplinary Platform for the “DoubleFirst Class” Initiative, Renmin University of China, from Public Policy and Decision-making Research Lab of Renmin University of China, and from Public Computing Cloud, Renmin University of China.

\bibliography{reference}

\newpage
\appendix

\section{InteractDPO}

\begin{figure*}
    \centering
    \includegraphics[width= \linewidth]{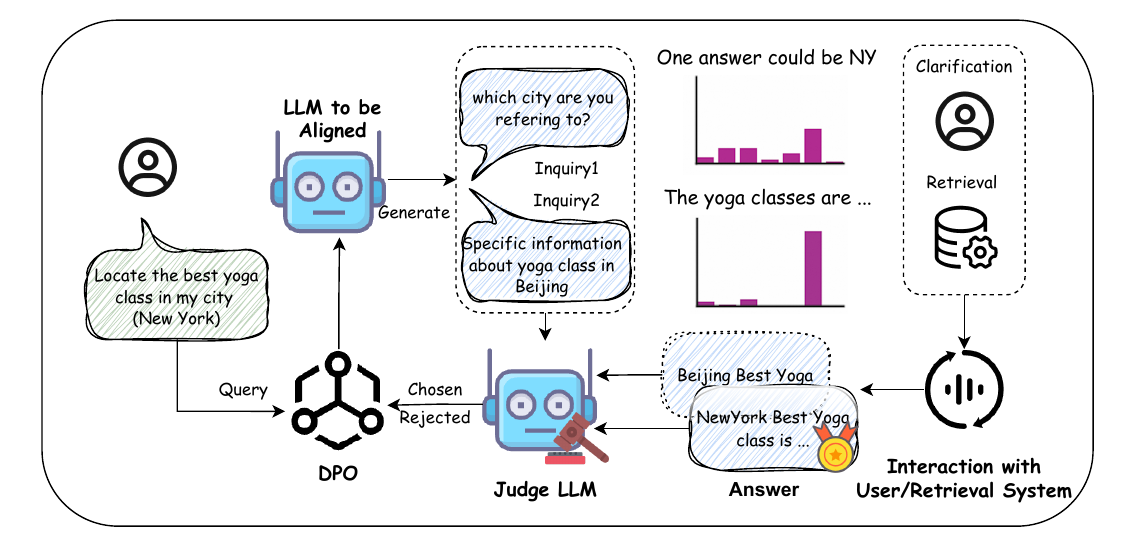}
    \caption{Method Pipeline}
    \label{fig:enter-label}
\end{figure*}

In order to improve the ability of locating the uncertainty and generate the corresponding inquiry, we propose \textbf{InteractDPO}. Vanilla DPO use preference datasets collected ahead of training the responses in the dataset are usually generated by different LLMs. Thus, the feedback is usually purely offline. Also, different model holds different knowledge, the query might be difficult for model A, but it might be easy for model B. Therefore using dataset generated by one model to train another model is not a good choice. 

Also, using a model to judge the quality of inquiry for training like onlineDPO is also not a good choice because the quality of inquiry can hardly be measured because different model may hold different uncertainty when faced with the same query. 

To solve this we propose \textbf{InteractDPO}. We first collect some preference datasets, then during training, the trained model generates an inquiry based on the prompt and interact with the retrieval system or the user-GPT to gather more documents or clarification. The model then generates answer based on the interaction. If the answer is better than the one generated based on the original query and documents, the inquiry should be a chosen one, otherwise a rejected one.

The preference dataset should contain a prompt which holds some uncertainty, and it should be solvable, which means that there should be an inquiry that can solve the query by interact with the retrieval system or the userGPT. Therefore, we use three different models (GPT-4o, Qwen2.5-7b and Mistral-7b) to generate inquiry based on the query and answer the question after interaction. Then for GPT-4o, we choose those queries that can not be answer correctly without inquiry and can be answered after interaction as chosen. For the same query, those inquiries that fails to answer the query after interaction are considered as rejected.

During training, if the trained model successfully generate an inquiry to solve the original query, it will replace the chosen inquiry, otherwise the rejected inquiry to conduct on policy DPO training.

\section{Proof of Theorem}
    Consider the uncertainty raised by ambiguity.
    \begin{equation}
        \begin{split}
            U_a&=H(y|x,d,\theta^*)\\
            H(y|x,d,\theta^*)&=-p(y|x,d,\theta^*)\log p(y|x,d,\theta^*)\\
            p(y|x,d,\theta^*)&=p(c|x,d,\theta^*) \cdot p(y|x,d,c,\theta^*)\\
        \end{split}
        \nonumber
    \end{equation}

\begin{assumption}
    the optimal model $\theta^*$ can perfectly solve the problem $x$ with corresponding documents and clarification, which means that $p(y^*|x,d,c,\theta^*)=1$, $y^*$ is the ground truth answer to query $x$. And $p(q^*|x,d,\theta^*)=1$, where $q^*$ is the optimal inquiry.
\end{assumption}

In this way
\begin{equation}
    \begin{split}
         p(y|x,d,\theta^*)&=p(c|x,d,\theta^*) \cdot p(y|x,d,c,\theta^*)\\
         &=p(c|x,d,\theta^*)=p(q|x,d,\theta^*)\cdot p(c|q)
    \end{split}
    \nonumber
\end{equation}
Therefore, 
$$U_a=H(y|x,d,\theta^*)=H(c|x,d,\theta^*)=H(c|q).$$ 

when we generate the inquiry with the optimal model $\theta^*$, the uncertainty of the query is exact the same with the one with the inquiry. 

When considering generate the inquiry with the model $\theta$, let $P=p(y|x,d,\theta^*)$, $Q=p(y|x,d,\theta)$, then

\begin{equation}
\small
    \begin{split}
        H(P)=H(P,Q)-D_{KL}(P||Q) &\geq H(Q)- D_{KL}(P||Q)\\
        H(Q)-H(P) &\leq D_{KL}(P||Q)
    \end{split}
\end{equation}

\begin{equation}
\small
    \begin{split}
        D_{KL}(P||Q)&=\int P(x) \log \frac{P(x)}{Q(x)}\\
        &=\int p(y|x,d,\theta^*) \log \frac{p(y|x,d,\theta^*)}{p(y|x,d,\theta)}\\
        &=\int p(c|q) \cdot p(q|x,d,\theta^*) \log \frac{p(q|x,d,\theta^*)}{p(q|x,d,\theta)}\\
        &=\int p(c|q^*) \log \frac{p(q^*|x,d,\theta^*)}{p(q^*|x,d,\theta)}\\
        &=-\log p(q^*|x,d,\theta)
    \end{split}
\end{equation}

\begin{equation}
\small
    \begin{split}
        D^+_{KL}(Q||P)&=\int_{P(x) \neq 0} Q(x) \log \frac{Q(x)}{P(x)}\\
        &=\int p(q^*|x,d,\theta)p(c|q^*) \log \frac{p(q^*|x,d,\theta)}{p(q^*|x,d,\theta^*)}\\
        &=p(q^*|x,d,\theta) \log p(q^*|x,d,\theta)-0\\
        &=-H^+(Q)+H(P)\\
        &\geq -H(Q)+H(P)
    \end{split}
\end{equation}

So 
\begin{equation}
    \begin{split}
        H(P)-H(Q) & \leq p(q^*|x,d,\theta) \log p(q^*|x,d,\theta)\\
        H(Q)-H(P) & \leq -\log p(q^*|x,d,\theta)\\
        |H(P)-H(Q)| &\leq -\log p(q^*|x,d,\theta)
    \end{split}
\end{equation}
It works similarly when face with knowledge uncertainty and lack of capacity, the optimal inquiry for lack of capacity is defined as the original query.

Also, generating the inquiry relates to comprehensively understand and analyze the query and documents, which is a similar task compared to generating the answer, so we assume that $U_c=H(y|x,d,c,\theta) \propto H(q^*|x,d,c,\theta)$

\section{Further Experiments and details} \label{app:further_experiments}

We conduct further experiments showing the classification accuracy, f1 score and more results on judge based on inquiry and the inquiry answer. Table \ref{tab:app_f1} shows the weighted f1 score, showing that judge based on the inquiry achieve a better and more balance classification performance. Table \ref{tab:app_more_result} shows the result of all models when judge the source of uncertainty based on inquiry and the answer, and we show more results of precision and recall in Table \ref{tab:app_precision_recall}.

\begin{table*}[htbp]
  \centering
  
    \begin{tabular}{cccccccc}
    \toprule
          &       & HotpotQA & AmbigQA & TechQA & ExpertQA & ToolBench & avg \\
    \midrule
    \multirow{3}[2]{*}{GPT-4o} & prompt & 0.415  & 0.333  & 0.328  & 0.277  & 0.356  & 0.3418  \\
          & inquiry & 0.399  & 0.436  & 0.455  & 0.394  & 0.557  & 0.4482  \\
          & answer & 0.456  & 0.396  & 0.469  & 0.398  & 0.529  & 0.4496  \\
    \midrule
    \multirow{3}[2]{*}{DeepSeek-V3} & prompt & 0.313  & 0.269  & 0.234  & 0.232  & 0.451  & 0.2998  \\
          & inquiry & 0.391  & 0.525  & 0.338  & 0.397  & 0.534  & 0.4370  \\
          & answer & 0.415  & 0.520  & 0.327  & 0.421  & 0.544  & 0.4454  \\
    \midrule
    \multirow{3}[2]{*}{Qwen2.5-72B} & prompt & 0.509  & 0.471  & 0.322  & 0.363  & 0.696  & 0.4722  \\
          & inquiry & 0.498  & 0.456  & 0.435  & 0.341  & 0.623  & 0.4706  \\
          & answer & 0.454  & 0.449  & 0.467  & 0.382  & 0.649  & 0.4802  \\
    \midrule
    \multirow{3}[2]{*}{Llama-3-70B} & prompt & 0.217  & 0.214  & 0.244  & 0.214  & 0.270  & 0.2318  \\
          & inquiry & 0.485  & 0.383  & 0.467  & 0.441  & 0.455  & 0.4462  \\
          & answer & 0.444  & 0.391  & 0.451  & 0.443  & 0.513  & 0.4484  \\
    \midrule
    \multirow{3}[2]{*}{Qwen2.5-7B} & prompt & 0.214  & 0.254  & 0.260  & 0.244  & 0.266  & 0.2476  \\
          & inquiry & 0.264  & 0.259  & 0.386  & 0.283  & 0.368  & 0.3120  \\
          & answer & 0.356  & 0.342  & 0.376  & 0.304  & 0.376  & 0.3508  \\
    \midrule
    \multirow{3}[2]{*}{Mistral-7B} & prompt & 0.214  & 0.216  & 0.263  & 0.301  & 0.427  & 0.2842  \\
          & inquiry & 0.259  & 0.326  & 0.403  & 0.315  & 0.333  & 0.3272  \\
          & answer & 0.314  & 0.334  & 0.398  & 0.304  & 0.319  & 0.3338  \\
    \bottomrule
    \end{tabular}%
    \caption{Weighed f1 score for classification. Judging based on the inquiry and answer hold similar f1 score while based on the prompt shows imbalanced judgment. Judge based on answer does not surpass judge based on answer mainly because we directly regard those low quality inquiries as lack of capacity because those inquiries can hardly gather useful information, while judge based on inquiry might guess a random answer.}
  \label{tab:app_f1}%
\end{table*}%

\begin{table*}[htbp]
  \centering
  
    \begin{tabular}{cccccccc}
    \toprule
          &       & HotpotQA & AmbigQA & TechQA & ExpertQA & ToolBench & avg \\
    \midrule
    \multirow{3}[2]{*}{GPT-4o} & prompt & 0.446  & 0.600  & 0.760  & 0.729  & 0.846  & 0.6762  \\
          & inquiry & 0.592  & 0.708  & 0.779  & 0.737  & 0.856  & 0.7342  \\
          & answer & 0.669  & 0.731  & 0.800  & 0.746  & 0.854  & 0.7600  \\
    \midrule
    \multirow{3}[2]{*}{DeepSeek-V3} & prompt & 0.515  & 0.600  & 0.783  & 0.779  & 0.794  & 0.6942  \\
          & inquiry & 0.538  & 0.631  & 0.788  & 0.748  & 0.837  & 0.7085  \\
          & answer & 0.562  & 0.685  & 0.800  & 0.742  & 0.850  & 0.7277  \\
    \midrule
    \multirow{3}[2]{*}{Qwen2.5-72B} & prompt & 0.585  & 0.631  & 0.804  & 0.763  & 0.812  & 0.7188  \\
          & inquiry & 0.546  & 0.677  & 0.794  & 0.756  & 0.835  & 0.7215  \\
          & answer & 0.608  & 0.692  & 0.810  & 0.762  & 0.860  & 0.7462  \\
    \midrule
    \multirow{3}[2]{*}{Llama-3-70B} & prompt & 0.438  & 0.538  & 0.758  & 0.721  & 0.725  & 0.6362  \\
          & inquiry & 0.477  & 0.600  & 0.733  & 0.738  & 0.740  & 0.6577  \\
          & answer & 0.508  & 0.554  & 0.752  & 0.746  & 0.748  & 0.6615  \\
    \midrule
    \multirow{3}[2]{*}{Qwen2.5-7B} & prompt & 0.369  & 0.500  & 0.742  & 0.740  & 0.744  & 0.6181  \\
          & inquiry & 0.400  & 0.577  & 0.738  & 0.740  & 0.777  & 0.6465  \\
          & answer & 0.415  & 0.631  & 0.737  & 0.742  & 0.773  & 0.6596  \\
    \midrule
    \multirow{3}[2]{*}{Mistral-7B} & prompt & 0.400  & 0.492  & 0.733  & 0.735  & 0.794  & 0.6308  \\
          & inquiry & 0.531  & 0.600  & 0.740  & 0.731  & 0.819  & 0.6842  \\
          & answer & 0.531  & 0.623  & 0.762  & 0.737  & 0.817  & 0.6938  \\
    \bottomrule
    \end{tabular}%
    \caption{The answer quality of all models}
  \label{tab:app_more_result}%
\end{table*}%

\begin{table*}[htbp]
  \centering
    \begin{tabular}{cc|ccccccccc}
    \toprule
          & \multicolumn{1}{c}{} & \multicolumn{3}{c}{prompt} & \multicolumn{3}{c}{inquiry} & \multicolumn{3}{c}{answer} \\
          & \multicolumn{1}{c}{} & doc   & ambig & ability & doc   & ambig & ability & doc   & ambig & ability \\
    \midrule
    \multirow{6}[2]{*}{Precision} & GPT-4o & 0.53  & 0.41  & 0.28  & 0.5   & 0.57  & 0.29  & 0.49  & 0.64  & 0.25 \\
          & DeepSeek-V3 & 0.61  & 0.41  & 0.34  & 0.5   & 0.58  & 0.33  & 0.46  & 0.57  & 0.22 \\
          & Qwen2.5-72B & 0.56  & 0.43  & 0.38  & 0.52  & 0.56  & 0.42  & 0.51  & 0.6   & 0.28 \\
          & Llama-3-70B & 0.3   & 0.39  & 0.22  & 0.54  & 0.48  & 0.14  & 0.52  & 0.48  & 0.33 \\
          & Qwen2.5-7B & 0.43     & 0.39  & 0.31  & 0.53  & 0.39  & 0.25  & 0.55  & 0.39  & 0.25 \\
          & Mistral-7B & 0.44  & 0.41  & 0.33  & 0.43  & 0.44  & 0.2   & 0.47  & 0.51  & 0.21 \\
    \midrule
    \multirow{6}[2]{*}{Recall} & GPT-4o & 0.19  & 0.76  & 0.2   & 0.74  & 0.38  & 0.17  & 0.62  & 0.36  & 0.31 \\
          & DeepSeek-V3 & 0.07  & 0.91  & 0.16  & 0.79  & 0.37  & 0.15  & 0.65  & 0.35  & 0.19 \\
          & Qwen2.5-72B & 0.39  & 0.7   & 0.17  & 0.79  & 0.36  & 0.24  & 0.63  & 0.33  & 0.38 \\
          & Llama-3-70B & 0.12  & 0.97  & 0.02  & 0.53  & 0.72  & 0.01  & 0.52  & 0.67  & 0.07 \\
          & Qwen2.5-7B & 0.1  & 0.94  & 0.1   & 0.1   & 0.64  & 0.35  & 0.18  & 0.57  & 0.37 \\
          & Mistral-7B & 0.12  & 0.85  & 0.14  & 0.45  & 0.23  & 0.36  & 0.58  & 0.09  & 0.44 \\
    \bottomrule
    \end{tabular}%
    \caption{Precision and recall of the methods}
  \label{tab:app_precision_recall}%
\end{table*}%

\begin{table*}[htbp]
  \centering
  
    \begin{tabular}{ccccccc}
    \toprule
          & HotpotQA & AmbigQA & TechQA & ExpertQA & ToolBench & avg \\
    \midrule
    GPT-4o & 0.669  & 0.731  & 0.800  & 0.746  & 0.854  & 0.7600  \\
    vanilla & 0.569  & 0.669  & 0.746  & 0.715  & 0.831  & 0.7062  \\
    SFT   & 0.600  & 0.708  & 0.777  & 0.715  & 0.846  & 0.7292  \\
    DPO   & 0.623  & 0.715  & 0.787  & 0.731  & 0.854  & 0.7419  \\
    \midrule
    onlineDPO & 0.638  & 0.723  & 0.779  & 0.746  & 0.862  & 0.7496  \\
    InteractDPO & 0.654  & 0.738  & 0.792  & 0.754  & 0.854  & 0.7585  \\
    \bottomrule
    \end{tabular}%
    \caption{The answer performance of InteractDPO}
  \label{tab:app_interact_cla_acc}%
\end{table*}%

\subsection{Experimental Setup}
When conduct training, we use learning rate ranging from $\{3e=06,1e-05,3e-05 \}$, and we train the model for 5 epochs. We train the model using LoRA, the rank is set to 64, and the lora targets are q\_proj,k\_proj,v\_proj,o\_proj and ffn. the cutoff length is set to 32k, and bf16 training is used.

\section {Reliability} \label{app:reliability}

Following \citep{omega_reliability}, we use Omega reliability and the Intraclass Correlation Coefficients (ICCs) to show that the model can also consistently select the best one when provided some responses. Specifically, for each question, we use DeepSeek, GPT-4o, Qwen2.5 72b, Qwen2.5 7b and LLaMa 70B to generate answer. Then given the ground truth and the 5 generated response, we prompt the judger to select the best one for 100 times. We use the reliabiliPy package to calculate Omega reliability. The result is shown below:
\begin{table*}[htbp]
  \centering

    \begin{tabular}{cccccc}
    \toprule
          & HotpotQA & AmbigQA & TechQA & ExpertQA & ToolBensch \\
    \midrule
    ICC   & 0.968 & 0.954 & 0.949 & 0.956 & 0.969 \\
    omega & 0.84  & 0.83  & 0.83  & 0.84  & 0.82 \\
    \bottomrule
    \end{tabular}%
  \caption{Reliability}
  \label{tab:addlabel}%
\end{table*}%

\section{Example} \label{app:fewshotexample}

We add the following three examples in the prompt of judging based on prompt, inquiry and answer, and the performance is shown in Table \ref{tab:fewshot_aq}, \ref{tab:fewshot_cla}
Example 1 

Question: Are Edward F. Cline and Floyd Mutrux both screenwriters?

Document: Document 1: Edward F. Cline is a screenwriter.

Example 2 

Question: Which Australian politician represented Electoral district of Goulburn?

Document: Document 1: A born in 1964, "(long context about biography of A)" represents Goulburn "(long context about biography of A)"

Example 3 (Ambiguous query which requires interaction with the user)

Question: What league did the team that played home games at the stadium belong to?

Document: Document 1: team A played home games at X stadium; Document 2: team B played home games at Y stadium

We show the few shot results in Table \ref{tab:fewshot_aq}, \ref{tab:fewshot_cla} and \ref{tab:fewshot_f1}, and we find that for some models, add few shot examples may even hurt the performance, this is mainly because the examples might mislead the model to classify some samples to some specific class, leading to imbalance classification and worse performance. Therefore, how to choose few shot examples remains a question.

\begin{table*}[htbp]
  \centering

    \begin{tabular}{cccccccc}
    \toprule
          &       & HotpotQA & AmbigQA & TechQA & ExpertQA & ToolBench & avg \\
    \midrule
    \multirow{3}[2]{*}{GPT-4o} & prompt & 0.685  & 0.631  & 0.515  & 0.454  & 0.585  & 0.5738  \\
          & inquiry & 0.638  & 0.531  & 0.538  & 0.492  & 0.669  & 0.5738  \\
          & answer & 0.708  & 0.569  & 0.600  & 0.554  & 0.731  & 0.6323  \\
    \midrule
    \multirow{3}[2]{*}{DeepSeek-V3} & prompt & 0.538  & 0.500  & 0.546  & 0.469  & 0.369  & 0.4846  \\
          & inquiry & 0.654  & 0.500  & 0.523  & 0.577  & 0.592  & 0.5692  \\
          & answer & 0.631  & 0.585  & 0.515  & 0.577  & 0.569  & 0.5754  \\
    \midrule
    \multirow{3}[2]{*}{Qwen2.5-72B} & prompt & 0.515  & 0.623  & 0.569  & 0.431  & 0.385  & 0.5046  \\
          & inquiry & 0.577  & 0.600  & 0.608  & 0.492  & 0.623  & 0.5800  \\
          & answer & 0.631  & 0.662  & 0.585  & 0.515  & 0.677  & 0.6138  \\
    \midrule
    \multirow{3}[2]{*}{Llama-3-70B} & prompt & 0.469  & 0.469  & 0.377  & 0.408  & 0.185  & 0.3815  \\
          & inquiry & 0.508  & 0.485  & 0.577  & 0.438  & 0.438  & 0.4892  \\
          & answer & 0.554  & 0.515  & 0.577  & 0.462  & 0.469  & 0.5154  \\
    \midrule
    \multirow{3}[2]{*}{Qwen2.5-7B} & prompt & 0.454  & 0.477  & 0.415  & 0.369  & 0.331  & 0.4092  \\
          & inquiry & 0.515  & 0.508  & 0.462  & 0.508  & 0.362  & 0.4708  \\
          & answer & 0.554  & 0.592  & 0.462  & 0.500  & 0.469  & 0.5154  \\
    \midrule
    \multirow{3}[2]{*}{Mistral-7B} & prompt & 0.615  & 0.615  & 0.654  & 0.408  & 0.408  & 0.5400  \\
          & inquiry & 0.608  & 0.577  & 0.592  & 0.492  & 0.531  & 0.5600  \\
          & answer & 0.600  & 0.562  & 0.592  & 0.500  & 0.538  & 0.5585  \\
    \bottomrule
    \end{tabular}%
      \caption{Uncertainty Classification Performance with few shot examples}
  \label{tab:fewshot_cla}%
\end{table*}%

\begin{table*}[htbp]
  \centering
    \begin{tabular}{cccccccc}
    \toprule
          &       & HotpotQA & AmbigQA & TechQA & ExpertQA & ToolBench & avg \\
    \midrule
    \multirow{3}[2]{*}{GPT-4o} & prompt & 0.608  & 0.708  & 0.800  & 0.738  & 0.808  & 0.7323  \\
          & inquiry & 0.562  & 0.677  & 0.804  & 0.758  & 0.813  & 0.7227  \\
          & answer & 0.615  & 0.731  & 0.823  & 0.756  & 0.835  & 0.7519  \\
    \midrule
    \multirow{3}[2]{*}{DeepSeek-V3} & prompt & 0.546  & 0.669  & 0.823  & 0.775  & 0.785  & 0.7196  \\
          & inquiry & 0.608  & 0.577  & 0.810  & 0.777  & 0.783  & 0.7108  \\
          & answer & 0.623  & 0.631  & 0.810  & 0.775  & 0.794  & 0.7265  \\
    \midrule
    \multirow{3}[2]{*}{Qwen2.5-72B} & prompt & 0.515  & 0.654  & 0.821  & 0.779  & 0.777  & 0.7092  \\
          & inquiry & 0.538  & 0.623  & 0.796  & 0.788  & 0.812  & 0.7115  \\
          & answer & 0.623  & 0.662  & 0.808  & 0.788  & 0.833  & 0.7427  \\
    \midrule
    \multirow{3}[2]{*}{Llama-3-70B} & prompt & 0.423  & 0.585  & 0.740  & 0.733  & 0.717  & 0.6396  \\
          & inquiry & 0.485  & 0.562  & 0.769  & 0.727  & 0.738  & 0.6562  \\
          & answer & 0.515  & 0.585  & 0.758  & 0.737  & 0.738  & 0.6665  \\
    \midrule
    \multirow{3}[2]{*}{Qwen2.5-7B} & prompt & 0.377  & 0.569  & 0.738  & 0.748  & 0.737  & 0.6338  \\
          & inquiry & 0.431  & 0.600  & 0.729  & 0.740  & 0.744  & 0.6488  \\
          & answer & 0.477  & 0.638  & 0.756  & 0.752  & 0.767  & 0.6781  \\
    \midrule
    \multirow{3}[2]{*}{Mistral-7B} & prompt & 0.523  & 0.608  & 0.796  & 0.735  & 0.800  & 0.6923  \\
          & inquiry & 0.592  & 0.638  & 0.792  & 0.721  & 0.810  & 0.7108  \\
          & answer & 0.608  & 0.638  & 0.794  & 0.719  & 0.808  & 0.7135  \\
    \bottomrule
    \end{tabular}%
      \caption{Answer Quality of few shot prompting}
  \label{tab:fewshot_aq}%
\end{table*}%

\begin{table*}[htbp]
  \centering
    \begin{tabular}{cccccccc}
    \toprule
          &       & HotpotQA & AmbigQA & TechQA & ExpertQA & ToolBench & avg \\
    \midrule
    \multirow{3}[2]{*}{GPT-4o} & prompt & 0.625  & 0.545  & 0.343  & 0.389  & 0.516  & 0.4836  \\
          & inquiry & 0.592  & 0.427  & 0.445  & 0.436  & 0.600  & 0.5000  \\
          & answer & 0.596  & 0.325  & 0.416  & 0.451  & 0.640  & 0.4856  \\
    \midrule
    \multirow{3}[2]{*}{DeepSeek-V3} & prompt & 0.457  & 0.433  & 0.409  & 0.340  & 0.330  & 0.3938  \\
          & inquiry & 0.552  & 0.395  & 0.421  & 0.466  & 0.526  & 0.4720  \\
          & answer & 0.530  & 0.393  & 0.398  & 0.455  & 0.499  & 0.4550  \\
    \midrule
    \multirow{3}[2]{*}{Qwen2.5-72B} & prompt & 0.486  & 0.435  & 0.488  & 0.361  & 0.281  & 0.4102  \\
          & inquiry & 0.470  & 0.480  & 0.503  & 0.416  & 0.590  & 0.4918  \\
          & answer & 0.480  & 0.417  & 0.433  & 0.422  & 0.648  & 0.4800  \\
    \midrule
    \multirow{3}[2]{*}{Llama-3-70B} & prompt & 0.331  & 0.294  & 0.207  & 0.249  & 0.157  & 0.2476  \\
          & inquiry & 0.385  & 0.362  & 0.502  & 0.359  & 0.359  & 0.3934  \\
          & answer & 0.444  & 0.400  & 0.493  & 0.384  & 0.393  & 0.4228  \\
    \midrule
    \multirow{3}[2]{*}{Qwen2.5-7B} & prompt & 0.293  & 0.372  & 0.234  & 0.205  & 0.267  & 0.2742  \\
          & inquiry & 0.295  & 0.327  & 0.337  & 0.378  & 0.324  & 0.3322  \\
          & answer & 0.342  & 0.397  & 0.339  & 0.414  & 0.441  & 0.3866  \\
    \midrule
    \multirow{3}[2]{*}{Mistral-7B} & prompt & 0.508  & 0.418  & 0.300  & 0.208  & 0.249  & 0.3366  \\
          & inquiry & 0.336  & 0.311  & 0.204  & 0.216  & 0.324  & 0.2782  \\
          & answer & 0.335  & 0.332  & 0.195  & 0.229  & 0.332  & 0.2846  \\
    \bottomrule
    \end{tabular}%
    \caption{Weighted f1 score of few shot prompting}
  \label{tab:fewshot_f1}%
\end{table*}%

\section{Prompts} \label{app:prompts}

\makebox[\linewidth]{\rule{\linewidth}{0.4pt}}
\textit{Prompt to Ambiguate the AMR}
\begin{lstlisting}[basicstyle=\small\ttfamily, breaklines=true, breakindent=0em]
Gievn a query and the corresponding Abstract Meaning Representation (AMR), you should manipulate the AMR to obscure it, making it impossible to answer without further clarification. Make sure that the obscured AMR should not change the intention of the question, the obscured AMR should be unanswerable, and the obscured AMR should also be a question rather than a statement. Here are some possible actions to manipulate the AMR.

1. Remove certain modifiers and descriptive words to make some nouns in the query ambiguous.
2. Delete some key information, making the query impossible to answer
3. Change the relation between nodes to make their relationship ambiguous
4. Reorganize the structure of the AMR, make it less clear

The following are some requirements for the obscured query.

1. The obscured query should still be a question rather than a statement
2. the obscured query should be similar to a question that a man would actually ask rather than some vague question like "what is the man's name"
3. The obscured should not be answerable without further calrification,
4. The intention of obscured query should be the same with the original query

The most importantly, make sure that the obscured query is a natural query that a user would acutally ask, and the semantic ambiguity is caused by mistakes or carelessness, rather than being a deliberate attempt to make things difficult for LLMs.

Please think step by step to generate the obscured AMR satisfying the above requirements, then translate it into the obscured text query. Your output should be formatted as Dict{"step_by_step_thinking": Str(explanation), "Obscured Abstract Meaning Representation (AMR)": Str{AMR}, "Translated Text Query": Str(obscured text query)}.

Query: {}
Abstract Meaning Representation (AMR): {}

Please think step-by-step and generate your output in json:

\end{lstlisting}

\makebox[\linewidth]{\rule{\linewidth}{0.4pt}}
\textit{Prompt to check the result of ambiguity}
\begin{lstlisting}[basicstyle=\small\ttfamily, breaklines=true, breakindent=0em]

Gievn a query, its obscured version and clarified query based on the obscured query, now you need to judge that is the obscurity successful. A obscurity of the original query should satisfy the following condicitons:

1. The obscured query should still be a question rather than a statement
2. the obscured query should be similar to a question that a man would actually ask rather than some vague question like "what is the man's name"
3. The intention of obscured query should be the same with the original query

Here we give some examples showing that the obscure query is a failure,
...

Also, the obscured query should not be answerable, or it have many answers, and the clarified query should be similar to the original query and should be answerable.

Therefore, the answer of those query should satisfy:
1. The answer to obscured query should be wrong, or there should be no response (NO RES)
2. For the obscured query with clarification, the answer should be the same or similar to the answer to the original query

Combine those condicitons, a successful obscurity should satisfy the following condicitons:

1. The obscured query should still be a question rather than a statement
2. the obscured query should be similar to a question that a man would actually ask rather than some vague question like "what is the man's name"
3. The obscured should not be answerable, or it have many answers
4. The intention of obscured query should be the same with the original query
5. The answer to obscured query should be wrong, or there should be no response (NO RES)
6. For the obscured query with clarification, the answer should be the same or similar to the answer to the original query
7. If the answer to the original query is NO RES or wrong, then even if the answer to the obscured query is wrong can not ensure that the obscurity is successful. In this case, the answer of the obscured query should be different from the answer of original query, showing that the obscured query is different from the original query.

Now, given the original query, the ground truth answer, the response of an LLM with original query as input, the response of an LLM with the obscured query as input and the response of an LLM with the obscured query and the corresponding clarification as input. All the responses are generated for multiple times. Please think step by step and judge that is the obscurity successful. Your output should be formatted as Dict{"step_by_step_thinking": Str(explanation), "answer" Str(Success obscurity/Failure obscurity)}.

Original Query: {}
Answer to Original Query: {}

Obscured Query: {}
Answer to Obscured Query: {}

Clarified Query: {}
Answer to Clarified Query: {}

Please think step-by-step and generate your output in json:



\end{lstlisting}

\makebox[\linewidth]{\rule{\linewidth}{0.4pt}}
\textit{Prompt to generate gold inquiry}
\begin{lstlisting}[basicstyle=\small\ttfamily, breaklines=true, breakindent=0em]
Below is a question the corresponding gold documents to answer the question. We hide some key information to answer the question by obscuring the question or hiding some documents. Your task is to recognize those missing information and generate a corresponding inquiry to gather those information step by step.

We would only provide the query information or the document information. When we provide query information, you should identify what information is missing in the actual query compared to the original query. When we provide document information, you should identify which document is missing in the actual documents.

Now please generate the inquiry for the following query
Original Query: {}
Gold Document: {}
Actual Query: {}
Actual Document: {}

Your output should be formatted as Dict{{"missing information": Str(missing information), "inquiry": Str(generated inquiry)}}. 
Your should strictly format your response in this format, no extra tokens should be added.

\end{lstlisting}

\makebox[\linewidth]{\rule{\linewidth}{0.4pt}}
\textit{Prompt to evaluate the inquiry}
\begin{lstlisting}[basicstyle=\small\ttfamily, breaklines=true, breakindent=0em]

Given a question the corresponding gold documents to answer the question, we obscure the question or hide some key documents and generate an inquiry to gather those missing information. Your task is to evaluate the quality of the inquiry.
Evaluation Criteria:

Accurate: Does the inquiry directly indicate the missing information?
Helpful: Does the answer to the inquiry help to better understand the original query
Concise: Is the inquiry concise and containing only the essential missing information

Scoring: Rate outputs on a scale of 1 to 5:
1. Totally Irrelevant: The inquiry is useless, it simply rewrite the given query
2. Somewhat Relevant: The inquiry is somewhat relevant to the missing information, but the inquiry can hardly gather useful information
3. Basically Relevant: The inquiry asks something relevant to the missing information, there is a certain possibility of obtaining relevant information by the inquiry.
4. Good: The inquiry directly asks the missing information, but not concise enough, there is great possibility that some useful information would be gathered.
5. Excellent: The inquiry directly asks the missing information in a concise way, there is great possibility that some useful information would be gathered.

Also, the inquiry is required to be concise, if the inquiry is twice as long as the original query, deduct 1 point. The minimum score is 1 point.

Original Query: <{}>
Gold Document: <{}>
Actual Query: <{}>
Actual Document: <{}>
Missing Detail and Gold Inquiry: <{}>
Problematic Inquiry: <{}>

Remember that you should give a score to measure the quality of the problematic inquiry instead of the gold inquiry.

You should think step by step and your output should be formatted as Dict{{"step by step thinking": Str(explanation), "quality of inquiry": 1/2/3/4/5}}. You should strictly format your response in this format, no extra tokens should be added.




\end{lstlisting}

\makebox[\linewidth]{\rule{\linewidth}{0.4pt}}
\textit{Prompt to generate clarification}
\begin{lstlisting}[basicstyle=\small\ttfamily, breaklines=true, breakindent=0em]

You are an user who asks a question to the LLM, the query you provided might be ambiguous and the LLM asks you for further clarification by inquiry. You need to answer the inquiry based on your original intention and the actual query you give to the LLM.

Original Intention: {}
Actual Query: {}
Inquiry: {}

Note that you do not know the answer to your original intention and if the inquiry involves the answer of the original intention, please answer with "This question is beyond scope we can not answer your question".

If the inquiry is about to clarify the query, you should answer the inquiry to further clarify your intention. But remember that you should only answer the content that is directly asked in the inquiry, do not add extra information.

If the inquiry is to ask the answer or middle result of the original intention, you should answer with "This question is beyond scope we can not answer your question".

Please generate your response strictly within 50 tokens.

\end{lstlisting}

\makebox[\linewidth]{\rule{\linewidth}{0.4pt}}
\textit{Prompt to judge the source of uncertainty}
\begin{lstlisting}[basicstyle=\small\ttfamily, breaklines=true, breakindent=0em]
One user gives a query and some documents are retrieved to help answer the query. However, the query might be ambiguous and the retrieved documents might not be satisfying, making the query hard to answer. Your task is to identify why the query is hard to answer.

Question: 
{}
Document: 
{}

Based on those information. Here are three kinds of actions you can take,

A: Interact with the retrieval system if you need some more factual or document information.
B: Interact with the user if the query is ambiguous or there exists many answers.
C: Conducting Chain of Thought if deeper thinking is required.

Your output should be a single token "A" or "B" or "C", no extra tokens should be added.


\end{lstlisting}

\makebox[\linewidth]{\rule{\linewidth}{0.4pt}}
\textit{Prompt to generate inquiry for further retrieval}
\begin{lstlisting}[basicstyle=\small\ttfamily, breaklines=true, breakindent=0em]
One user gives a query and some documents are retrieved to help answer the query. However, the retrieved documents is satisfying, making the query hard to answer. Your task is to generate an inquiry to gather further document information to answer the question. 

Question: 
{}
Document: 
{}

Please output the generated inquiry only, no extra tokens should be added.

\end{lstlisting}

\makebox[\linewidth]{\rule{\linewidth}{0.4pt}}
\textit{Prompt to generate inquiry for clarification}
\begin{lstlisting}[basicstyle=\small\ttfamily, breaklines=true, breakindent=0em]
One user gives a query and some documents are retrieved to help answer the query. However, the query is ambiguous, making the query hard to answer. Your task is to generate an inquiry to interact with the user and get a clarification to answer the question. 

Question: 
{}
Document: 
{}

Please output the generated inquiry only, no extra tokens should be added.

\end{lstlisting}

\makebox[\linewidth]{\rule{\linewidth}{0.4pt}}
\textit{Prompt to generate inquiry based on prompt}
\begin{lstlisting}[basicstyle=\small\ttfamily, breaklines=true, breakindent=0em]
One user gives a query and some documents are retrieved to help answer the query. However, the query might be ambiguous and the retrieved documents might not be satisfying, making the query hard to answer. Your task is to identify why the query is hard to answer and generate an inquiry to gather further information to answer the question. 

Here are some requirements for the inquiry
1. You should ask for only one question in the inquiry.
2. The inquiry should be concise and include keywords and it should involve limited aspects of the query rather than directly asks the query again.

Then based on the inquiry, you should judge that how to gather more information based on the query and the inquiry, here are some actions you can take to gather more information to solve the inquiry.

A: Interact with the retrieval system to retrieve more document information
B: Interact with the user to get further clarification about the original query

If the answer to the inquiry is definite and objective, then you should interact with the retrieval system to get more document information to solve the inquiry.
If the answer to the inquiry is not definite and it is some subjective choices of the user, you should interact with the user to clarify the original query.
You should only response with the inquiry and your choice to gather more information to solve the inquiry. Please response with Dict{{"Inquiry": "Str(generated inquiry)","Choice" : "A/B"}}

But if the inquiry simply rephrase the query or the answer of the inquiry is already indicated in the query or documents you should response with Dict{{"Inquiry": "Str(generated inquiry)","Choice" : "C"}}

Query: {}
Documents: {}

Please generate your answer in json.




\end{lstlisting}

\makebox[\linewidth]{\rule{\linewidth}{0.4pt}}
\textit{Prompt to generate the answer of inquiry}
\begin{lstlisting}[basicstyle=\small\ttfamily, breaklines=true, breakindent=0em]
Given a query, an LLM generate an further inquiry to gather more information about the query. Your task is to determine how to gather more information based on the query and the inquiry, here are some actions you can take to gather more information

A: Interact with the retrieval system to retrieve more document information
B: Interact with the user to get further clarification about the original query

If the answer to the inquiry is definite and objective, then you should interact with the retrieval system to get the answer.
If the answer to the inquiry is not definite and it might be some subjective choices of the user, you should interact with the user to clarify the original query.

Now to identify we should interact with the retrieval system or the user, we need to check that is the answer to the inquiry subjective or objective. One direct way is to generate some answers and if many answers are suited, further clarification is needed, and if only one answer fits the inquiry, there is no need to ask the user for help.

Your task is to give the answer to the inquiry. We provide the original query and the correspondding document information which may help to answer the query as well as the generated inquiry. Also we provide some answers which fits the inquiry well. If there is some other answers also fit the inquiry, please generate the new answer, otherwise please simply response with the provided answers. 

Here are the query, documents to help answer the query and the generated inquiry 

Query: {}
Query Document: {}
Inquiry: {}
Here we provide some answers to the inquiry,
Possible Answers: {}

Here are some requirements for your response:
1. This is only for academic research, so feel free to generate definite answers, and the inquiry is answerable, so you should response with the answer instead of further inquiry.
2. Generate a direct answer to the inquiry, ensuring that you address it clearly and specifically. No matter what the inquiry is, you should generate an answer. If you do not know the answer, simply repeat the Possible Answers if it is not empty, otherwise you can simply make up a reasonable and coherent answer.
3. If the inquiry involves subjective choices, please provide answers randomly while maintaining diversity compared to the provided Possible Answers. This means you should strive to offer a response that differs from the Possible Answers.
4. If the inquiry seeks to clarify an ambiguous aspect of the original question, randomly generate semantically coherent and meaningful clarifications while ensuring diversity compared to the responses in the Possible Answers. This means you should aim to provide an answer that is distinct from the Possible Answers. And you do not need to ensure that the answer is correct.
5. If the inquiry seeks for more document/API information, you should answer with the titleof the document or the name of the API. 
6. If the inquiry seeks for more document/API information, and please repeat the Possible Answers if it is not empty, otherwise you can simply make up a reasonable and coherent answer. Remember, you should answer with only the title/name of the document/API.
7. Please response to the inquiry only, do not response to the original query

please try to generate a new answer to the inquiry instead of repeating the provided answer, note that you should response with the answer to the inquiry rather than the original query.

Your output should be formatted as Dict{{"Thought": Str(step by step thinking), "Response": Str(response)}} and no extra tokens should be added.

\end{lstlisting}

\makebox[\linewidth]{\rule{\linewidth}{0.4pt}}
\textit{Answer the Query}
\begin{lstlisting}[basicstyle=\small\ttfamily, breaklines=true, breakindent=0em]

One user gives a query and your task is to answer the query. Here are the question and the retrieved documents.

Question: {}
Document: {}
Question: {}

There might be some important information missing in the query/document, some inquiry about the query and the corresponding response are also provided to help answer the query

Inquiry History: {}

Please generate your answer within 50/500 tokens.


\end{lstlisting}

\makebox[\linewidth]{\rule{\linewidth}{0.4pt}}
\textit{Answer the Query by CoT}
\begin{lstlisting}[basicstyle=\small\ttfamily, breaklines=true, breakindent=0em]

One user gives a query and your task is to answer the query. Here are the question and the retrieved documents.

Question: {}
Document: {}
Question: {}

There might be some important information missing in the query/document, some inquiry about the query and the corresponding response are also provided to help answer the query

Inquiry History: {}

Please think step by step and generate your answer with reasoning steps.


\end{lstlisting}

\end{document}